\documentclass[journal, 10pt, twocolumn]{IEEEtran}
\usepackage{graphicx, amsfonts}
\usepackage{subfigure}
\usepackage{float}
\usepackage{multirow}
\usepackage{tabularx}
\usepackage{makecell}
\usepackage{booktabs}
\usepackage{dsfont}
\usepackage{amsmath}
\usepackage{amssymb}
\usepackage{threeparttable}
\usepackage{color} %,hyperref}
\usepackage[justification=centering]{caption}
\usepackage{epstopdf}
\usepackage[lined,boxed,commentsnumbered, ruled]{algorithm2e}
\usepackage[normalem]{ulem}
\usepackage{verbatim}
\newcommand{\citep}{\cite}

\begin{document}
\title{Neighborhood Preserved Sparse Representation for Robust Classification on Symmetric Positive Definite Matrices}
\author{Ming Yin, ~Shengli Xie, ~Yi Guo, ~Junbin Gao and Yun Zhang % <-this % stops a space
\IEEEcompsocitemizethanks{\IEEEcompsocthanksitem Ming Yin, Shengli Xie and Yun Zhang are with School of Automation, Guangdong University of Technology, Guangzhou, 510006, China.
E-mail: yiming@gdut.edu.cn, shlxie@gdut.edu.cn, zy@gdut.edu.cn \protect
\IEEEcompsocthanksitem Yi Guo is Commonwealth Scientific and Industrial Research Organisation, North Ryde, NSW 1670, Australia. E-mail: yi.guo@csiro.au\protect
\IEEEcompsocthanksitem Junbin Gao is with The University of Sydney Business School, The University of Sydney, Camperdown, NSW 2006, Australia.
E-mail: junbin.gao@sydney.edu.au.}
\thanks{Manuscript received xxx; revised xxx.}}
% \markboth{IEEE Transactions on NEURAL NETWORKS AND LEARNING SYSTEMS}
% {Yin \MakeLowercase{\textit{et al.}}: Riemanninan SRC}

\maketitle
\begin{abstract}
Due to its promising classification performance, sparse representation based classification(SRC) algorithm has attracted great attention in the past few years. However, the existing SRC type methods apply only to vector data in Euclidean space. As such, there is still no satisfactory approach to conduct classification task for symmetric positive definite (SPD) matrices which is very useful in computer vision. To address this problem, in this paper, a neighborhood preserved kernel SRC method is proposed on SPD manifolds. Specifically, by embedding the SPD matrices into a Reproducing Kernel Hilbert Space (RKHS), the proposed method can perform classification on SPD manifolds through an appropriate Log-Euclidean kernel. Through exploiting the geodesic distance between SPD matrices, our method can effectively characterize the intrinsic local Riemannian geometry within data so as to well unravel the underlying sub-manifold structure. Despite its simplicity, experimental results on several famous database demonstrate that the proposed method achieves better classification results than the state-of-the-art approaches.
\end{abstract}
\IEEEpeerreviewmaketitle

\section{Introduction}
%\IEEEPARstart{I}{n}
In the past a few years, inspired by advances in $\ell_0$-norm and $\ell_1$-norm techniques, sparse representation\citep{DonohoEladTemlyakov2006} has been widely applied in computer vision, such as image segmentation\citep{YangWrightMaSastry2008}, image deblurring\citep{YinGaoTienCai2014} and face recognition\citep{WrightYangGaneshSastryMa2009}. It is worth noting that Wright \emph{et al}. proposed a sparse representation based classification(SRC) \citep{WrightYangGaneshSastryMa2009} to classify facial images, which is the first time to exploit the discriminative nature of sparse represeor face recognition.
In fact, facial images have a high dimensionality, which usually lie on a low-dimensional subspace or sub-manifold. Thus, Yang \emph{et al}. \citep{YangChuZhangXuYang2013} proposed a novel dimensionality reduction method that adopts SRC as a criterion to steer the design of a feature extraction method. In addition, many high-dimensional data in real world may be better modeled by nonlinear manifolds. To overcome the nonlinear obstruction, some researches suggest to map these data into a kernel feature space by using some nonlinear mapping, and then SRC is performed in this new feature space by utilizing kernel trick \citep{ZhangZhouChangLiuYanWangLi2012}\citep{YinLiuJinYang2012}.

However, most of the above work mainly focuses on the problem associated with vector-valued data. The higher-order signals like images (2D, 3D or higher) have to be dealt with primarily by vectorizing them and applying any of the available vector techniques. As a result, such type vector features cannot efficiently characterize the high-dimensional data in computer vision, machine learning and medical image analysis\citep{YinGaoGuo2015}. Concretely, in traditional sparse representation based classification, the sparsity representation for each query image is attained by a dictionary composed of all gallery data across all classes in a \emph{linear combination} way. Recent advance\citep{TuzelPorikliMeer2008} suggests that encoding images through symmetric positive definite (SPD) matrices and then interpreting such matrices as points on Riemannian manifolds can lead to promising classification performance. For instance, the human facial images are regarded as samples from a nonlinear sub-manifold\citep{WangShanChenGao2008}. Unfortunately, the \emph{linear combination} is not applicable to this case where data may be better modeled by nonlinear manifolds \citep{JayasumanaHartleySalzmannLiHarandi2013}\citep{HarandiHartleyLovellSanderson2014}. In other words, the direct applications of linear combination model to matrix-valued data will result in the comprised performance as inaccurate representation. Consequently, the traditional SRC is also no longer available to classification on SPD matrices as points on Riemannian manifolds.

To address this problem, a few solutions have been recently proposed to generalize sparse coding problems to Riemannian manifolds, such as \citep{CherianSra2014}\citep{HoXieVemuri2013}\citep{SivalingamBoleyMorellasPapanikolopoulos2014}. The most common approach is to calculate the tangent space to the manifold at the mean of the data points so as to obtain a Euclidean approximation of the manifold\citep{TuzelPorikliMeer2008}. Inspired by this idea, Ho \emph{et al}. \citep{HoXieVemuri2013} firstly proposed a nonlinear generalization of sparse coding to handle the non-linearity of Riemannian manifolds, via flattening a SPD manifold using a fixed tangent space. In order to further measure the representation error effectively, in \citep{SivalingamBoleyMorellasPapanikolopoulos2014}, a tensor sparse coding framework was proposed for positive definite matrices based on the log-determinant divergence (Burg loss). Instead of using extrinsic similarity measures as work\citep{SivalingamBoleyMorellasPapanikolopoulos2014}, the authors\citep{CherianSra2014} proposed to use the intrinsic Riemannian distance on the manifold of SPD matrices. Although locally flattening Riemannian manifolds via tangent spaces can handle their non-linearity, it inevitably leads to very demanding computation due to switching back and forth between tangent spaces and the manifold\citep{HarandiHartleyLovellSanderson2014}. Furthermore, linear reconstruction of SPD matrices is not as natural as in Euclidean space and this may incur errors \citep{LiWangZuoZhang2013}. On the other line, to address this nonlinear problems via LRR, a nonlinear LRR model is proposed to extend the traditional LRR from Euclidean space to Stiefel manifold \cite{YinGaoGuo2015}, SPD manifold \cite{FuGaoHongTien2015} and abstract Grassmann manifold \citep{WangHuGaoSunYin2015} respectively. Low-rank representation based method, however, often suffers from high computational complexities as the nuclear norm regularized optimizing. From this view point, sparse representation based method can readily reduce the computational complexities greatly due to only solving $\ell_0$-norm optimization problems rather than nuclear-norm ones.

The existing sparse representation methods on SPD matrices can been shown to be effective for classification\citep{HarandiHartleyLovellSanderson2014}\citep{LiWangZuoZhang2013}, however, there still remain questions about classification in the multiple sub-manifolds setting\citep{WangSlavakisLerman2015} with sparse representation. As SPD matrices are often that low-dimensional data embedded in high-dimensional non-Euclidean spaces, their underlying sub-manifolds are geodesic and referred to Riemannian multi-manifolds. Let $X$ be a SPD matrix and hence a point on $\mathcal{S}_d^+$, it can be assumed residing on the tubular neighborhood of some unknown geodesic sub-manifold $\mathcal{M}_k (1 \leq k \leq K)$, of a Riemannian manifold. As for this issue, sparse representation based classification \citep{HarandiHartleyLovellSanderson2014}\citep{LiWangZuoZhang2013} has not been sufficiently explored yet. Another reason, may not trivial, the sparse coding coefficients may vary a lot even for similar query samples in classification task as the mechanism of $\ell_1$-minimization. As a result, the unsatisfied recognition rate will be achieved. Motivated by these observations, in this paper, we propose a neighborhood preserved sparse representation for robust classification on SPD matrices. When encoding the query sample, we aim to use the training samples lying in its vicinity as the training samples and the query sample may reside in the same sub-manifold leading to a better classification performance. Despite its simplicity, the proposed method performs well for classification task. Specifically, to thoroughly exploit the intrinsic geometry among data on Riemannian manifold, a neighborhood preserved prior induced from the geodesic distance, besides the sparsity, is imposed on the sparse coefficients so that the similar query data produce similar sparse codes.

The main contributions in our paper are summarized below.
\begin{enumerate}
\item To our best knowledge, it is the first attempts to formulate the local consistency into the sparse coding paradigm over a Riemannian manifold via embedding them into RKHS. It is significantly different from the work in \citep{ZhangZhouChangLiuYanWangLi2012}\citep{YinLiuJinYang2012} as the latter did not consider the Riemannian geometry structure within data.
\item To efficiently measure the neighborhood between data points on Riemannian manifold, we compare the two geodesic distance under Stein metric and Log-Euclidean metric, respectively. To our best knowledge, this is one of the first attempts, from the weighted structured perspective, to compare the benefit from this two metrics for analyzing SPD matrices.
\item We apply our proposed methods to several classification tasks where the data are depicted as region covariance matrices.
\end{enumerate}

The remainder of this paper is organized as follows. In Section
\ref{SectionII}, we give a brief review on the related works. Section
\ref{SectionIII} is dedicated to introducing our novel neighborhood preserved kernel SRC, termed as NPKSRC. Section \ref{SectionIV} presents experimental results on image classification tasks. Finally, Section \ref{SectionV} concludes our paper and also provides the directions for future improvements.

\section{Related Work}\label{SectionII}
Before we introduce our model, in this section, we briefly review the recent development of sparse representation based classification methods\citep{WrightYangGaneshSastryMa2009}\citep{HoXieVemuri2013}
and the analysis of Riemannian geometry of SPD manifold\citep{PennecFillardAyache2006}. For convenience, Table \ref{Notation} gives the notation used throughout this paper.

%\section*{Notation}
\begin{table}
\begin{center}
\caption{Notation used in this paper}
\begin{tabular}{ll}
\hline
\textbf{Notation}&  \textbf{Description} \\
\hline
$X$& data matrix \\
$\mathcal{X}$& 3D matrix or 3-order tensor \\
$\textbf{x}$& column vector\\
${x}_i$& the element at position $i$ of vector $\textbf{x}$\\
$X_{ij}$& the $(i,j)$-th entry of matrix $X$\\
$\| \textbf{x} \|_1$ & $\ell_1$ norm of  $\textbf{x} $\\
$\|\textbf{x} \|_2$&  $\ell_2$ norm of  $\textbf{x} $\\
$T$&  transpose operator\\
$ \left\| \cdot \right\|_F $&  matrix Frobenius norm defined as $\left\| X \right\|_F^2 = \sum\limits_{i } {\sum\limits_{j} {\left| {X_{ij} } \right|^2 }}$ \\
$\|X\|_*$& nuclear norm of $X$ defined by the sum of its singular values\\
$\textrm{tr}( \cdot )$& matrix trace operator\\
$\mathcal{S}_d^+$& space of $d \times d$ SPD matrices\\
Log map & principal matrix logarithm \\
$\mathbf{log}_X(\cdot)$& Log map from SPD manifold to a tangent space at $X$\\
$T_X \mathcal{S}_d^+$& tangent space at a point $X$ on $\mathcal{S}_d^+$, \\
& which is a vector space including the tangent vectors\\
& of all possible curves passing over $X$. \\
\hline
\end{tabular}
\end{center}
\end{table}\label{Notation}

\subsection{Classification via Sparse Representation}
Sparse representation based classification(SRC) has been well-known as its robustness to face recognition\citep{WrightYangGaneshSastryMa2009}. Suppose that there exist $n$ classes and $m_i$ training data for each class $i$. We denote by $Y_i$ the collection of training data in the $i$-th class and $Y= [Y_1, Y_2,..., Y_n]\in \mathcal{R}^{d \times N}, N= \sum_{i=1}^n{ m_i }$ by the collection of all training data over all  classes. Given a test sample $\mathbf{x} \in \mathcal{R}^d$, which belongs to one of the $n$ classes, the goal of SRC is to find out the class to which $\mathbf{x}$ belongs, by seeking its sparsest representation over all training data.

Concretely, the SRC solves the following optimization problem.
\[
\mathop{\min}\limits_{\mathbf{c}} ~\frac1{2}\|\mathbf{x} - Y\mathbf{c} \|_2^2 + \lambda \| \mathbf{c} \|_1.
\]
Once the problem is solved, the class of given test sample can be found as the class that best represents it using the corresponding training data in class-wise way. That is, let $\mathbf{c}_i^*$ be a vector whose only nonzero entries are the entries in $\mathbf{c}_i$ that are associated with class $i$, we can adopt the following rule to determine $\mathbf{x}$ as class $j$ that has the minimum residual.
\begin{align}
\text{label}(\mathbf{x})= \mathop{\arg\min}\limits_{j} ~\frac1{2}\|\mathbf{x} - Y_j\mathbf{c}_j^* \|_2^2.\label{rule}
\end{align}

\subsection{Riemannian Geometry on SPD Matrices}
In general, SPD matrices lie on a non-flat Riemannian manifold, whose structure is suitably characterized by the geodesic distance induced by Riemannian metric. That is, a natural way to measure closeness of data on a Riemannian manifold is geodesics, eg. curves analogous to straight lines in $\mathds{R}^n$. For any two data points on a manifold, geodesic distance is the length of the shortest curve on the manifold connecting them. For this reason, there are, currently, two popular distance measures in $\mathcal{S}_d^+$. One is the affine invariant Riemannian metric (AIRM) and the other is Log-Euclidean metric.

As one of true metrics of geodesic distance, AIRM is probably the most widely used Riemannian metric defined as follows\citep{PennecFillardAyache2006}. Given $X \in \mathcal{S}_d^+$, the AIRM of two tangent vectors $\mathbf{v}, \mathbf{w} \in T_X \mathcal{S}_d^+$ is defined as
\begin{align*}
&\langle \mathbf{v}, \mathbf{w}  \rangle = \langle X^{-1/2}\mathbf{v} X^{-1/2}, X^{-1/2} \mathbf{w} X^{-1/2} \rangle \\
&= \textrm{tr} ( X^{-1} \mathbf{v}  X^{-1} \mathbf{w} ).
\end{align*}
The geodesic distance between points $X, Y \in \mathcal{S}_d^+$  induced from AIRM is then
\begin{align}
&\delta_g (X, Y) = \|\mathbf{log} (X^{-1/2}Y X^{-1/2}) \|_F . \label{gAIRM}
\end{align}

However, the above distance induced by AIRM is computationally intensive resulting in a significant numerical burden. To overcome this drawback of AIRM, Log-Euclidean metric is defined on the Lie group of SPD matrices corresponding to a Euclidean metric in the logarithmic domain. Specifically, the distance under Log-Euclidean metric is denoted by,
\begin{align}
\delta_l(X,Y) =  \|\mathbf{log}(X)- \mathbf{log}(Y)\|_F. \label{gLE}
\end{align}

\subsection{Spare Representation on SPD Matrices}
Since SPD matrices belong to a Lie group which is a Riemannian manifold \citep{ArsignyFillardPennecAyache2007}, it cripples many methods that rely on linear reconstruction. Generally, there are two methods to deal with the non-linearity of Riemannian manifolds. One is to locally flatten the manifold to tangent spaces\citep{TuzelPorikliMeer2008}. The underlying idea is to exploit the geometry of the manifold directly. The other is to map the data into a feature space usually a Hilbert space \citep{JayasumanaHartleySalzmannLiHarandi2013}. Precisely, it is to project the data into RKHS through kernel mapping \citep{HarandiSandersonHartleyLovell2012}. Both of these methods are seeking a transformation so that the linearity re-emerges.

A typical example of the former method is the one in \citep{HoXieVemuri2013}. Let $X$ be a SPD matrix and hence a point on $\mathcal{S}_d^+$. $\mathcal{D} = \{D_1, D_2, ..., D_N \}, D_i \in \mathcal{S}_d^+$ is a dictionary. An optimization problem for sparse coding of  $X$ on a manifold $\mathcal{M}$ is formulated as follows
\begin{align}
\mathop{\min}\limits_{\mathbf{c}} ~\|\mathbf{w}\|_1 + \lambda \left\| \sum\limits_{i = 1}^N {c}_{i}\mathbf{log}_{X}(D_i)  \right\|_{X}^2, ~\textrm{ s.t. } \sum\limits_{i = 1}^N c_{i}=1,\label{SRManifold}
\end{align}
where $\mathbf{log}_X(\cdot)$ denotes Log map from SPD manifold to a tangent space at $X$,  $\mathbf{c} = [c_1, c_2,..., c_N]$ is the sparse vector and $\|\cdot\|_X$ is the norm associated with $T_X \mathcal{S}_d^+$. Because $\mathbf{log}_X(X)=\textbf{0}$, the second term in Eq.\eqref{SRManifold} is essentially the error of linearly reconstructing $\mathbf{log}_X(X)$ by others on the tangent space of $X$. %Since this tangent space is a vector space, this reconstruction is well defined. As a result, the traditional sparse representation model can be performed on Riemannian manifold.

Although locally flattening Riemannian manifolds via tangent spaces\citep{HoXieVemuri2013} can handle their non-linearity, it inevitably leads to very demanding computation due to
switching back and forth between tangent spaces and the manifold. Furthermore, linear reconstruction of SPD matrices is not as natural as in Euclidean space and this may incur errors. Thus, the kernel-based sparse coding on SPD matrices is proposed as follows\citep{HarandiSalzmann2015}.
\begin{align}
\mathop{\min}\limits_{\mathbf{c}} ~\|\mathbf{c}\|_1 + \lambda \left\| \phi(X)-  \sum\limits_{i = 1}^N {c}_{i}\phi(D_i)  \right\|_F^2, ~\textrm{ s.t. } \sum\limits_{i = 1}^N c_{i}=1,\label{KSRManifold}
\end{align}
where $\phi( \cdot )$ denotes a feature mapping function that projects SPD matrices into RKHS such that $\langle \phi(X),\phi(Y) \rangle = \kappa(X,Y)$ where $\kappa(X,Y)$ is a positive definite (PD) kernel.

\section{Kernel Sparse Representation on SPD Matrices via Neighborhood Preserved}\label{SectionIII}
However, the constraint of $\ell_1$-norm sparsity is beneficial to classification task, a test input might be reconstructed by training images, i.e., codewords, which are far away from the test sample\citep{WangYangYuLvHuangGong2010}. As a consequence, the SRC type methods will produce unsatisfying classification results. In addition, data of SPD matrices are often modeled as a union of low-dimensional sub-manifolds\citep{WangSlavakisLerman2015}. Under this context, classification algorithms aim at partitioning data based on the underlying low-dimensional  non-Euclidean spaces. Therefore, the neighborhood of each data on Riemannian manifold can be fit by a geodesic sub-manifold model.

Motivated by the above issues, in this section, we propose a neighborhood preserved kernel sparse representation based classification (termed as NPKSRC) algorithm on SPD matrices, by considering the structure within data points. The formulation can be written as following.
\begin{align}
\mathop{\min}\limits_{\mathbf{c}} ~\|\mathbf{w} \odot \mathbf{c}\|_1 + \lambda \left\| \phi(X)-  \sum\limits_{i = 1}^N {c}_{i}\phi(D_i)  \right\|_F^2, ~\textrm{ s.t. } \sum\limits_{i = 1}^N c_{i}=1,\label{NPKSRC}
\end{align}
where $\odot$ means element-wise multiplication and $\mathbf{w}$ is a vector imposing restriction on the structure of the solution. Similar to the prior work\citep{HarandiSalzmann2015}\citep{HoXieVemuri2013}, the affine constraint is applied to our model too. Furthermore, by introducing structure constraint, i.e., $\mathbf{w}$, we actually enforce a smaller weight on the samples belonging to the same sub-manifold with the test input, and vice versa. To some extent, the entries of $\mathbf{w}$ are denoting the affinity between the test input and the training data. Then, how to choose a informative entries of $\mathbf{w}$ is a key factor of success for the subsequent classification tasks.

\subsection{Analysis of the Weight Matrix}
The structure of data are often determined by using pairwise distance between data points \citep{WangYangYuLvHuangGong2010}. Moreover, manifold learning (neighborhood preservation model) has been widely used for dimension reduction by learning and embedding local consistency
of original data into a low-dimensional representation\citep{RoweisSaul2000}\citep{TenenbaumSilvaLangford2000}.
For simplicity, we assume only there exist a two-class data underlying geodesic sub-manifolds $\mathcal{S}_1$ and $\mathcal{S}_2$, respectively. Given a test input $X \in \mathcal{S}_d^+$, in general, there is a larger probability to assign it to that class determined by points lying on the sub-manifolds $\mathcal{S}_1$ if the nearby points of $X$ is that points located on $\mathcal{S}_1$. From this intuition, we can use the affinity, i.e., geodesic distance, between test input and training data to compute the weight. Concretely, $\mathbf{w}$ is constructed in terms of the geodesic distance of $X$ from every training sample(a subset of $\mathcal{D}$). As such, a locally smooth sparse code vector is achieved where the sparsity is a result of the neighborhood preserving since the training samples far away from $X$ do not contribute to its reconstruction. Therefore, in this paper, we utilize the geodesic distance, under Log-Euclidean metric, between a test input and training samples as the weight, illustrated as following.
\[ {w}_{i} = \delta_l(Y_i,X)=  \|\mathbf{log}(Y_i)- \mathbf{log}(X)\|_F.
\]

\subsection{The Proposed Method}
Given the training data $\mathcal{Y}= [Y_1,Y_2,..., Y_N]$ on SPD manifold, the corresponding kernel sparse representation algorithm is formulated as following.
\begin{align}
\mathop{\min}\limits_{\mathbf{c}} ~ \|\text{diag}(\mathbf{w})\mathbf{c}\|_1 + \frac{\lambda}{2} \left\| \phi(X)-  \sum\limits_{i = 1}^N {c}_{i}\phi(Y_i)  \right\|_2^2, ~\textrm{ s.t. } \sum\limits_{i = 1}^N c_{i}=1,\label{NPKSRC2}
\end{align}

Through expanding the $\ell_2$-norm term and some algebra manipulations, we will consider the following problem that has a same solution to problem \ref{NPKSRC2}. For clarity and completeness, the detailed derivation of the problem \ref{NPKSRC2} can be found in the appendix.
\begin{align*}
\mathop{\min}\limits_{\mathbf{c}}  \|\text{diag}(\mathbf{w}) \mathbf{c}\|_1+ \frac{\lambda}{2}
\| \mathbf{\bar{x}}- \bar{D} \mathbf{c}\|_2^2, ~\textrm{ s.t. } \sum\limits_{i = 1}^N c_{i}=1.
\end{align*}
where $\mathbf{\bar{x}}= \Sigma^{-1/2}U^T \kappa(X, \mathcal{Y})$ and $\bar{D} = \Sigma^{-1/2}U^T$, given the \text{SVD} of $\kappa(\mathcal{Y}, \mathcal{Y})$ is $U\Sigma U^T$.

Here, we adopt Log-Euclidean Gaussian kernel\citep{LiWangZuoZhang2013} to transform the SPD matrices into RKHS such that the linear combination will make sense. In contrast, the Log-Euclidean kernel can well characterize the true geodesic distance between SPD matrices instead. Specifically, a Log-Euclidean Gaussian kernel is defined by $ \kappa_g(X,Y) = \textrm{exp}\{-\gamma \|\mathbf{log}(X)- \mathbf{log}(Y)\|_F^2  \}$, which is a \textit{p.d.} kernel for any  $\gamma > 0$.

\subsection{Optimization}\label{opt}
To solve the problem\eqref{NPKSRC2}, we apply the well-known alternating direction method of multipliers (ADMM)\citep{BoydParikhChuPeleatoEckstein2011} here. Before directly using ADMM, we should decouple the variables in the problem \eqref{NPKSRC2} firstly. Let $W = \text{diag}(\mathbf{w})$ and introduce a variable $\mathbf{a}= W\mathbf{c}$. Then,
\begin{align}
&\mathop{\min}\limits_{\mathbf{a},\mathbf{c}}  \|\mathbf{a}\|_1+ \frac{\lambda}{2}
\| \mathbf{\bar{x}}- \bar{D} \mathbf{c}\|_2^2, \notag \\
& \quad\textrm{s.t.,} ~\mathbf{a}= W\mathbf{c}, ~{\textbf{c}}^T\textbf{1} = 1. \label{NPKSRC3}
\end{align}
where $\mathbf{1} \in \mathcal{R}^N$ is a column vector whose entries are all ones.

The above problem is not convex in both, however, it is convex in a variable for fixed another unrelated one. Hence, the augmented Lagrangian function of problem \eqref{NPKSRC3} can be written as follows.
\begin{align}
\mathcal{L}(\mathbf{a},\mathbf{c}) = \mathop{\min}\limits_{\mathbf{a},\mathbf{c}} \|\mathbf{a}\|_1+  \frac{\lambda}{2}
\| \mathbf{\bar{x}}- \bar{D} \mathbf{c}\|_2^2 \notag \\ +\frac{\mu}{2} (\| \mathbf{a}- W\mathbf{c}\|_2^2 + ({\textbf{c}}^T\textbf{1}- 1)^2) \notag \\
+ \Delta^T (\mathbf{a}- W\mathbf{c})+ \delta({\textbf{c}}^T\textbf{1}- 1). \label{NPKSRC4}
\end{align}

Thus, we optimize the problem by alternatively fixing other unrelated variables as follows.
\begin{enumerate}
\item Update $\mathbf{c}$,\\
\begin{align}
\mathop{\min}\limits_{\mathbf{c}}  \frac{\lambda}{2}
\| \mathbf{\bar{x}}- \bar{D} \mathbf{c}\|_2^2+\frac{\mu}{2}(\| \mathbf{a}- W\mathbf{c}\|_2^2 + ({\textbf{c}}^T\textbf{1}- 1)^2) \notag \\
+ \Delta^T (\mathbf{a}- W\mathbf{c})+ \delta({\textbf{c}}^T\textbf{1}- 1). \label{updateC}
\end{align}
Setting the derivative w.r.t. $\mathbf{c}$ to be zero gives the following.
\begin{align*}
0 = -\lambda\bar{D}^T(\mathbf{\bar{x}}- \bar{D} \mathbf{c})+ \mu W^T(W\textbf{c}- \textbf{a})\\
+\mu \mathds{1}_{N \times N}\mathbf{c}- \mu \mathbf{1}
-W^T\Delta + \delta \mathbf{1}.
\end{align*}
where $\mathds{1}_{N \times N}$ is the matrix of size ${N \times N}$ with all ones.

Then, \\
\begin{align}
&\mathbf{c}_k=(\lambda\bar{D}^T\bar{D}+ \mu W^TW + \mu \mathds{1}_{N \times N})^{-1}(\lambda\bar{D}^T\mathbf{\bar{x}} \notag \\
&  + \mu W \textbf{a} + (\mu -\delta )\mathbf{1} + W^T\Delta).\label{Update_c}
\end{align}

\item Update $\mathbf{a}$, \\ % by fixing $\mathbf{c}_k$ ,\\
\begin{align}
\mathop{\min}\limits_{\mathbf{a}} \|\mathbf{a}\|_1+ \frac{\mu}{2}\| \mathbf{a}- W\mathbf{c}_k\|_2^2 +\Delta^T (\mathbf{a}- W\mathbf{c}_k). \label{updateA}
\end{align}

That is, \\
\begin{align*}
\mathop{\min}\limits_{\mathbf{a}} \|\mathbf{a}\|_1+ \frac{\mu}{2} \| \mathbf{a}- (W\mathbf{c}_k- \frac{\Delta}{\mu}) \|_2^2.
\end{align*}
The above problem has the following closed-form solution given by shrinkage operator\citep{LinChenMa2009}. That is,
\begin{align}
\mathbf{a}_k = \mathcal{S}_{\frac1{\mu}}(W\mathbf{c}_k- \frac{\Delta}{\mu})
\end{align}
where $\mathcal{S}_{\eta}(\cdot)$ is a shrinkage operator acting on each element of the given matrix, and is defined as $\mathcal{S}_{\eta}(v) = \textrm{sgn}(v)\textrm{max}(|v|- \eta, 0) $.

\item Update $\Delta$ and $\delta$.
\begin{align}
\Delta_{k} = \Delta_{k-1} + \mu (\mathbf{a}_k- W\mathbf{c}_k).\notag \\
\delta_k = \delta_{k-1} + \mu ({\textbf{c}_k}^T\textbf{1}- 1).
\end{align}
\end{enumerate}
These iterative steps will be terminated when $\|\textbf{c}_k- \textbf{c}_{k-1}\|_{\infty} \leq \epsilon$ and $\| {\textbf{c}_k}^T\textbf{1}- 1\|_{\infty}  \leq \epsilon $ are satisfied.

\subsection{Classification}
Once the new representation of the test input is obtained, the decision rule \eqref{rule} is applied to determine its class finally.
The detailed procedure of classification using neighborhood preserved kernel sparse representation is described in Algorithm\ref{Alg1}.
\begin{algorithm}
\caption{Classification Using Neighborhood Preserved Kernel Sparse Representation on SPD matrices}
\SetKwData{Index}{Index}
\KwIn{Training data $\mathcal{Y}= [Y_1,Y_2,..., Y_N], Y_i \in \mathcal{S}_d^+$ sorted according to the label of each data point; A test sample $X \in \mathcal{S}_d^+$; $\lambda$ and $\mu$.}
\BlankLine
\textbf{Steps:}
\begin{enumerate}
\item Construct the weight vector $\mathbf{w}$ by calculating the geodesic distance between $X$ and $\mathcal{Y}$ in terms of equation \eqref{gLE}.
\item Solve \eqref{NPKSRC2} by ADMM explained in Section \ref{opt},
and obtain the optimal solution $ \mathbf{c}^* $.
\item Compute the residuals of the test sample $X$ over all classes and assign its label finally.
\end{enumerate}
\KwOut{the label of test sample.}
\label{Alg1}
\end{algorithm}

\subsection{Complexity Analysis and Convergence}
As for the computational cost of the proposed algorithm, it is mainly determined by the steps in ADMM. The total complexity of NPKSRC is, as a function of the number of data points, $\mathcal{O}(N^3+t N^2)$ where $t$ is the total number of iterations. The soft thresholding to update the sparse matrix $\textrm{C}$ in each step is relatively cheaper, much less than $\mathcal{O}(N^2)$.  For updating $\mathbf{c}$ we can pre-compute the Cholesky decomposition of $(\lambda\bar{D}^T\bar{D}+ \mu W^TW + \mu \mathds{1}_{N \times N})^{-1}$ at the cost of less than $\mathcal{O}(\frac12 N^3)$, then compute new $\mathbf{c}$ by using \eqref{Update_c} which has a complexity of $\mathcal{O}(N^2)$ in general.

The above proposed ADMM iterative procedure to the augmented Lagrangian problem \eqref{NPKSRC4} satisfies the general condition for the convergence theorem in \citep{LinLiuLi2015}.

\section{Experimental Results} \label{SectionIV}
\begin{figure}[ht]
\centering
  \subfigure[]{\includegraphics[width=0.23 \textwidth]{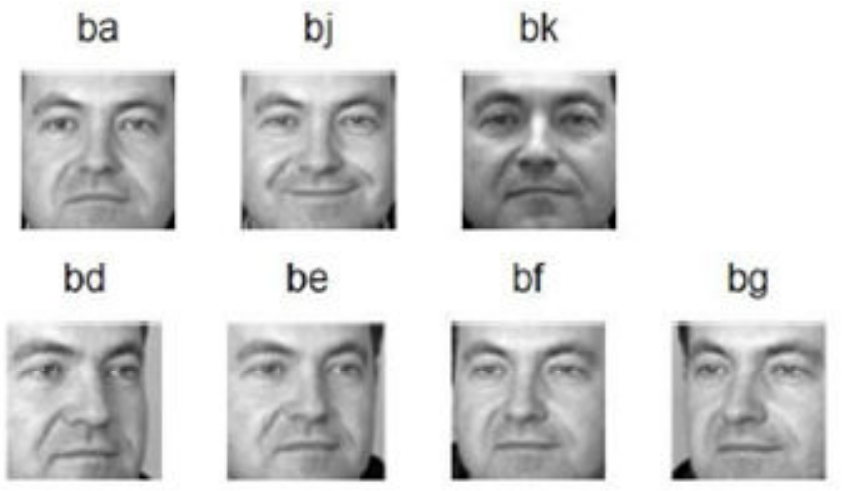}}
   \subfigure[]{\includegraphics[width=0.12 \textwidth]{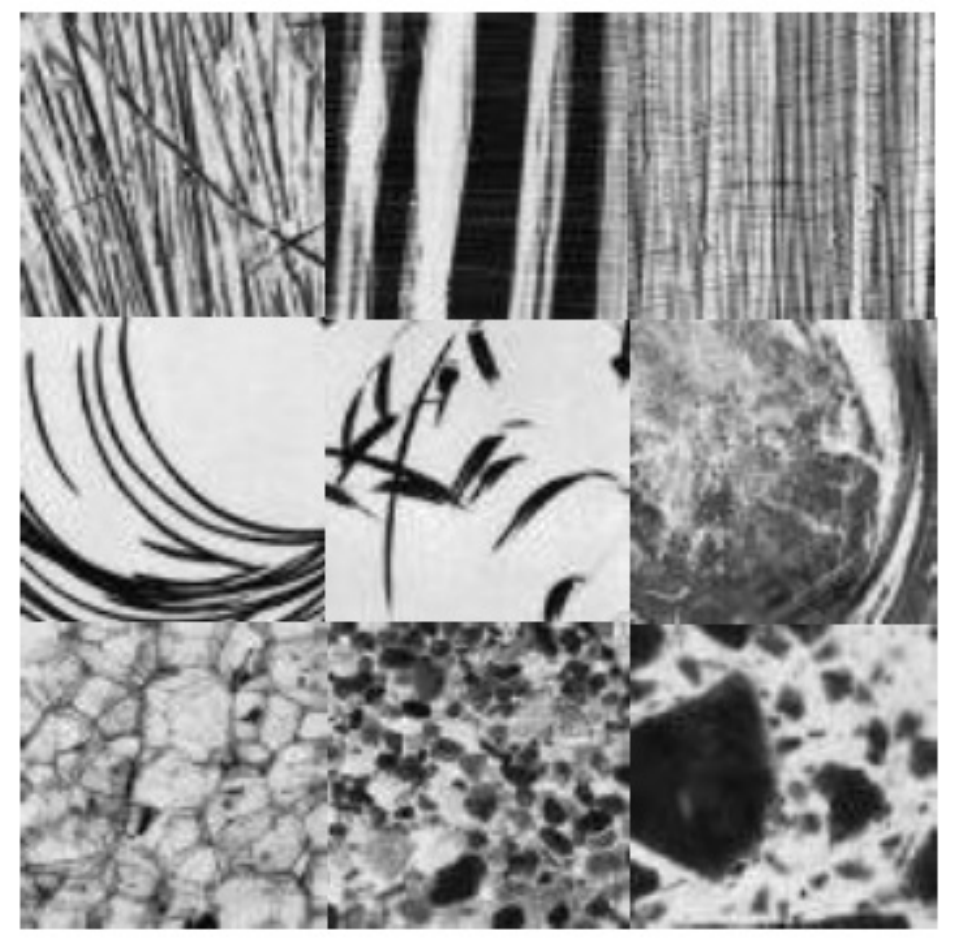}}
   \caption{Samples on the FERET (a) and Brodatz (b) database. }\label{fig1}
\end{figure}

In this section, we present several experimental results to demonstrate the effectiveness of NPKSRC. To comprehensively evaluate the performance of NPKSRC, we tested it on texture images, human faces and  pedestrian re-identification. Some sample images from test databases are shown in Figure \ref{fig1}.

We compare our proposed method with five state-of-the-art methods in terms of recognition accuracy.
\begin{enumerate}
\item  Sparse representation classification(SRC)\citep{WrightYangGaneshSastryMa2009} ;
\item  Gabor feature-based sparse representation in Euclidean space (GSRC)\citep{YangZhangShiuZhang2013} ;
\item  Classification using Riemannian sparse representation based on Riemannian distance  (RSRC)\citep{CherianSra2014};
\item  Classification using Riemnnian sparse representation based on Stein kernel (RSRS)\citep{HarandiHartleyLovellSanderson2014};
\item  Log-Euclidean Gaussian kernel sparse representation based classification (LogE-GkSRC) \citep{LiWangZuoZhang2013}.
\end{enumerate}

\subsection{Texture Classification}
Firstly, we used Brodatz texture database to conduct classification task.
In this dataset, it includes 5-texture (`5c', `5m', `5v', `5v2',`5v3'), 10-texture (`10', `10v') and 16-texture (`16c', `16v') mosaics. Before using the proposed method, we downsampled each image into 256 $\times$ 256 and then split into 64 regions of size 32 $\times$ 32. To obtain their Region Covariance Matrices (RCM), a feature vector $f(x,y)$ for any pixel $I(x,y)$ is extracted, e.g., $f(x,y) = (I(x,y), |\frac{\partial I}{\partial x}|, |\frac{\partial I}{\partial y}|, |\frac{\partial^2 I}{\partial x^2}|, |\frac{\partial^2 I}{\partial y^2}|)$. Then, each region can be depicted by a 5 $\times$ 5 covariance descriptor. As for the obtained RCM, there are 64 covariance matrices in each class. We randomly selected 5 from each class as training samples and the rest as the query samples. That means almost 8\% samples are selected as training data and the rest for testing classification. To achieve a stable result, the reported classification rate is averaged over 20 trials.

How to construct the weight vector $\mathbf{w}$ is not a trivial work in our method. Here, to better characterize the local geometry within data, we test the two kinds of distance by different metrics, i.e., Stein metric and Log-Euclidean metric. Brodatz-`16v' is selected as test dataset for classification task, which includes 16 classes. The classification results are shown in Table \ref{Tab2}. As can be seen, the geodesic distance under Log-Euclidean metric can better characterize the manifold structure of data by achieving a better classification rate.
\begin{table}
\caption{classification results in terms of accuracy (\%) on Brodatz-`16v' with different geodesic distance.}
\begin{center}
\begin{tabular}{|c|c|c| }
\hline
$Metric$ & Stein metric & LogE metric  \\
\hline
Accuracy &78.5  &79.34 \\
\hline
\end{tabular}
\end{center}
\label{Tab2}
\end{table}

\begin{figure}[ht]
\centering
  \subfigure[]{\includegraphics[width=0.45 \textwidth]{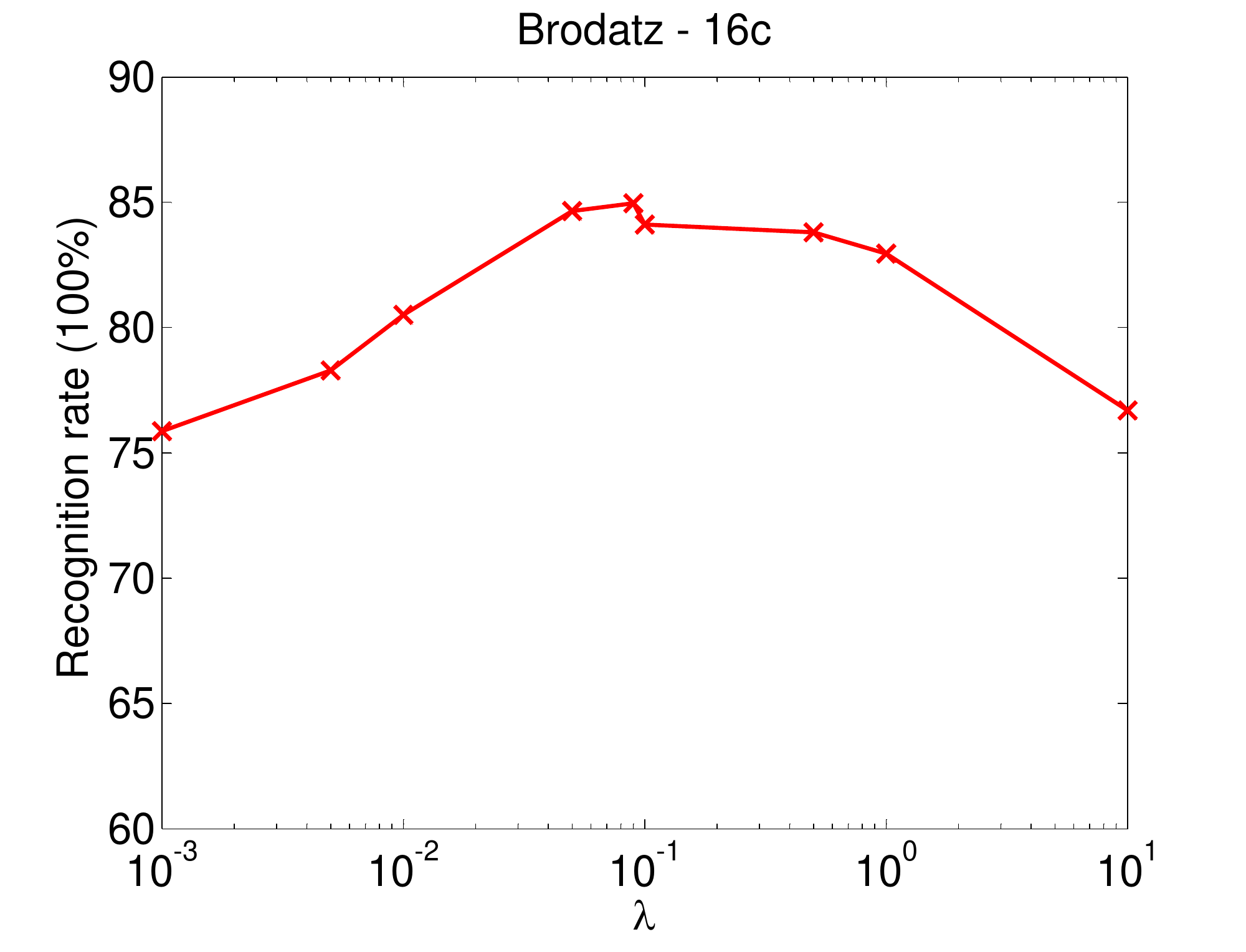}}
   \subfigure[]{\includegraphics[width=0.45 \textwidth]{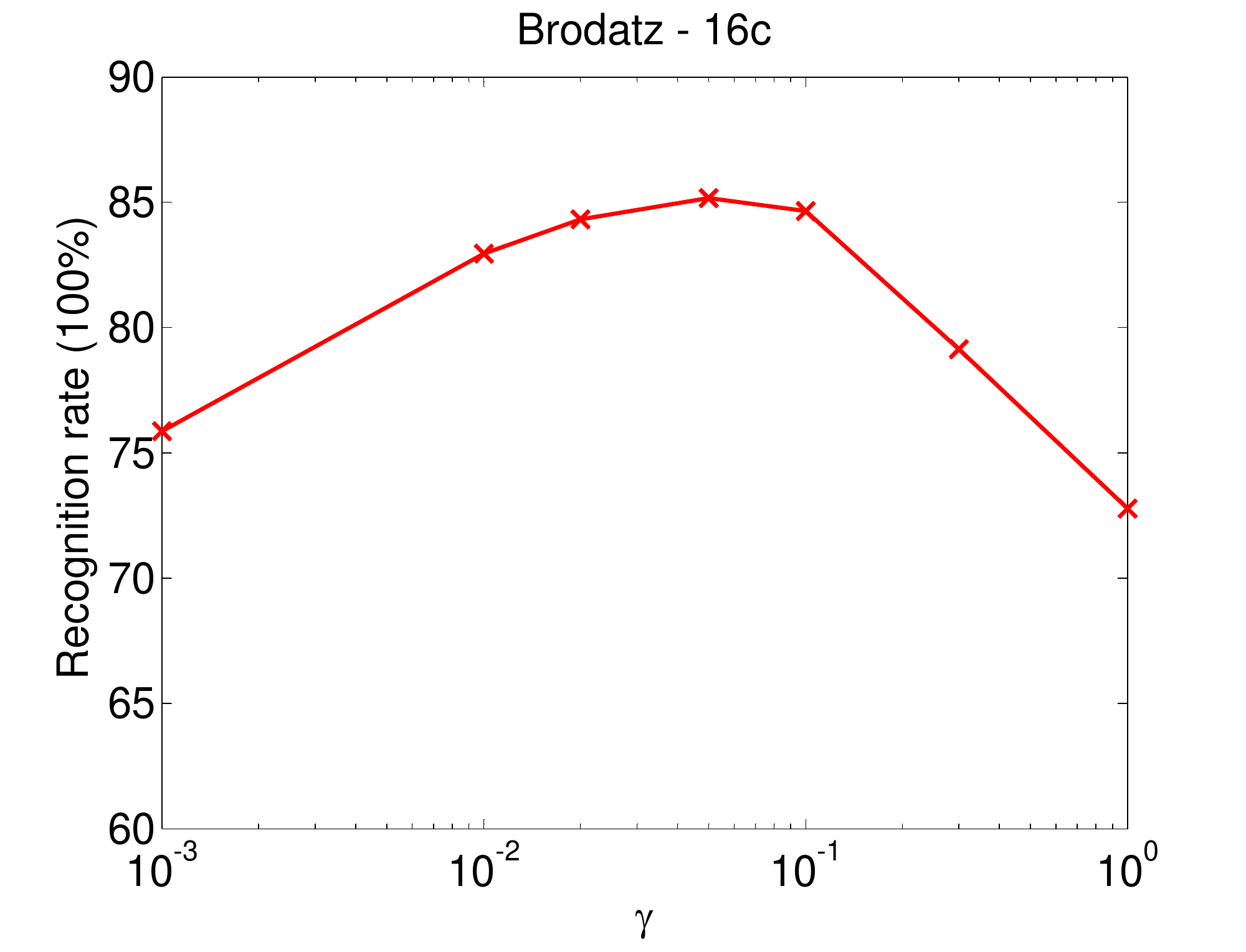}}
   \caption{Classification rate on Brodatz-16c  vs. parameter $\lambda$ and $\gamma$. }\label{fig2}
\end{figure}

To efficiently determine the parameters in our method, in Fig.\ref{fig2},we report the recognition accuracy on Brodatz-`16c' with varying parameters $\lambda$ and $\gamma$, respectively. From the figs., we can set $\lambda =0.09$ and $\gamma =0.05$ for the best recognition result.

By applying the different methods, we presented the classification results in Table \ref{Tab3}.  As well, the tuned parameters are reported for the results achieved by other methods. The bold numbers highlight the best results. From the results, we can observe that the proposed approach outperforms other methods in most cases while, on average, RSRS achieved the second best performance based on Stein kernel. This can be interpreted by the Stein distance may better suit some subset of Brodatz dataset, i.e., `5v2' and`5v3'. As for SRC, it conducts classification as a baseline due to the lack of consideration of the intrinsic geometry structure within data.

\begin{table*}
\caption{Classification results in terms of accuracy (\%) on Brodatz dataset.}\label{Tab3}
\begin{center}
\begin{tabular}{|l|c|c|c|c|c|c|c|c|c|c| }
\hline
$Dataset$ & `5c'& `5m'& `5v'& `5v2'&`5v3' &`16v' &`16c' &`10v' &`10' & avg. \\
\hline\hline
SRC\citep{WrightYangGaneshSastryMa2009} (0.001)&20.00 &20.00 &20.00 &20.00 &20.00 & 6.25 &44.07 &10.00 &20.00 &20.04 \\
%GSRC\cite{YangZhangShiuZhang2013} & &  &  &  &  &  &  &  & \\
RSRC\citep{CherianSra2014}(0.01) &97.97 &50.51 &83.05 &83.05 &71.19 &61.65 &74.05	&85.08 &92.88 &77.71\\
RSRS\citep{HarandiHartleyLovellSanderson2014}(10.0) &98.31& 89.15 & 83.05 &\textbf{87.12} &\textbf{88.14}& 72.46 & 83.79 & 88.31 & 94.24  &87.17\\
LogE-GkSRC\citep{LiWangZuoZhang2013}(0.001,0.02)  &97.29	&94.92 &83.73 &86.10 &86.78 &73.20 &80.08 &90.34	&94.58 &84.55 \\
NPKSRC(0.09,0.05) &\textbf{98.31} &\textbf{98.98} &\textbf{84.41}	&84.07 &87.12 &\textbf{79.34} &\textbf{88.35}&\textbf{91.02} &\textbf{98.31} &\textbf{89.26} \\
\hline
\end{tabular}
\end{center}
\end{table*}

\subsection{Face Recognition}
Next, we selected the ``b" subset of FERET database to further evaluate the classification performance, in which it covers 1400 images with the size of 80 $\times$ 80 from 200 subjects (about 7 each). This subset consists of the images, under different expression and illumination conditions, marked by `ba', `bd', `be', `bf', `bg', `bj', and `bk'. Specifically, training images include neutral expression `ba', smiling expression `bj', and illumination changes `bk', while test samples involve face images of varying pose angle such as `bd'+25$^{\circ}$, `be'+15$^{\circ}$, `bf'-15$^{\circ}$, and `bg'-25$^{\circ}$.

\begin{figure}[ht]
\centering
{\includegraphics[width=0.45 \textwidth]{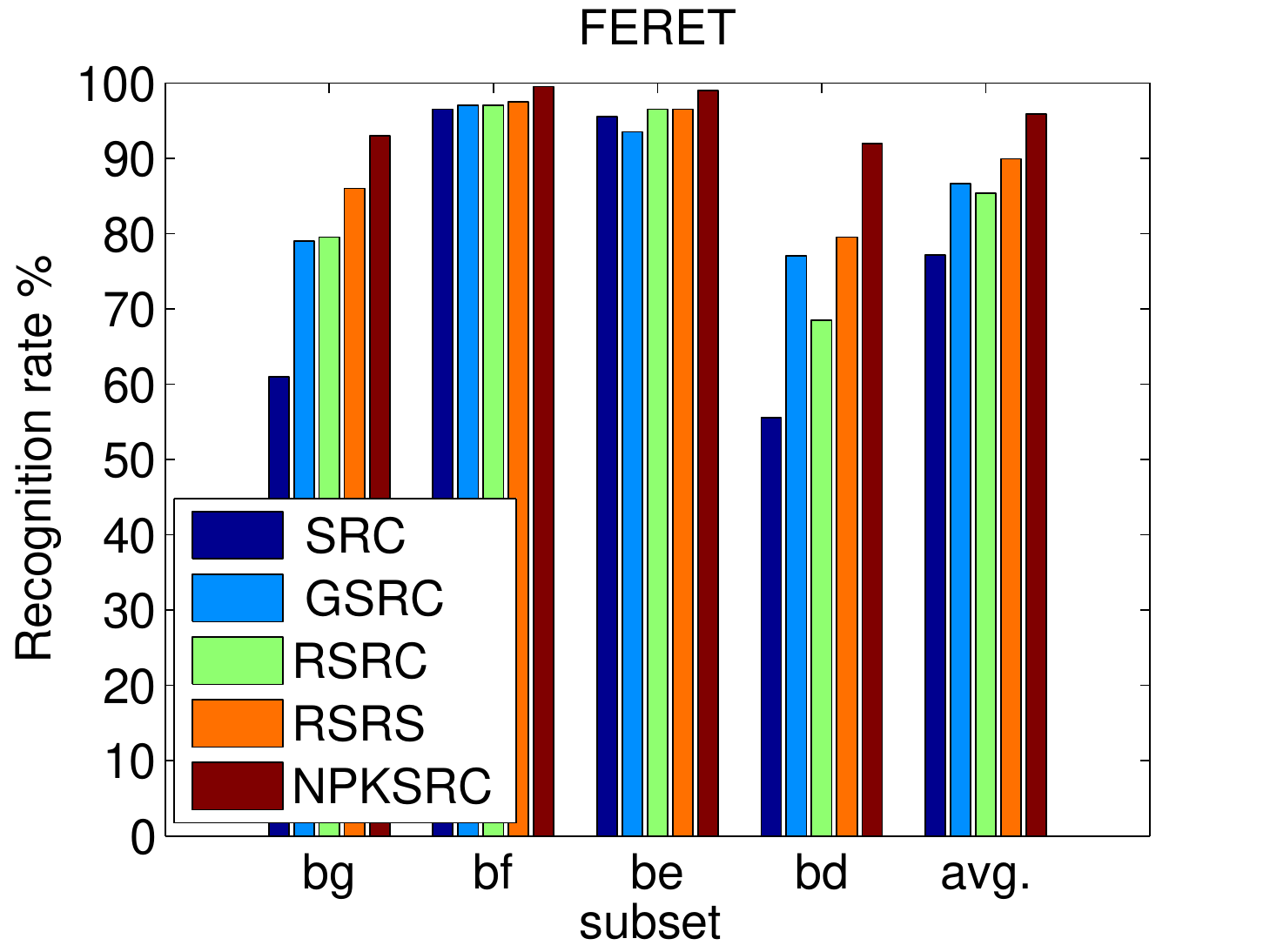}}
   \caption{Comparison of Recognition rate (\%) on FERET dataset. }\label{fig3}
\end{figure}

To represent a facial image, similar to the work \citep{YangZhangShiuZhang2013}, we created a 43$\times$43 region covariance matrix, i.e., a specific SPD matrix, which is composed of intensity value, spatial coordinates, 40 Gabor filters at 8 orientations and 5 scales. The down-sampling factor in Gabor filtering is applied too. For SRC, the Gabor features are firstly vectorized and the common SRC classifier is applied. The classification results achieved by other methods are reported in Table \ref{Tab4} and Fig.\ref{fig3}. The tuned parameters are presented in the table too. For NPKSRC, the $\lambda $ and $\gamma $ are set 0.9 and 0.02, respectively. From the table, we can see our proposed method achieves pleasing recognition performance compared to others. This is owed to the consideration of locality structure between data by using Riemannian metric.

\begin{table}
\caption{Classification results in terms of accuracy (\%) on FERET dataset.}\label{Tab4}
\begin{center}
\begin{tabular}{|l|c|c|c|c|c| }
\hline
$Dataset$ & `bg' &`bf' &`be' &`bd' & avg.\\
\hline\hline
SRC\citep{WrightYangGaneshSastryMa2009}(0.01)&61.00 &96.50 &95.50 &55.50 &77.13\\
GSRC\cite{YangZhangShiuZhang2013}(0.001) & 79.00 & 97.00 & 93.50 & 77.00 & 86.60 \\
RSRC\citep{CherianSra2014} (0.01) & 79.50 &97.00 &96.50 &68.50 &85.38  \\
RSRS\citep{HarandiHartleyLovellSanderson2014}(10.0) &86.00 &97.50  &96.50  &79.50  &89.90   \\
NPKSRC (0.9, 0.02) &\textbf{93.00} &\textbf{99.50} &\textbf{99.00} &\textbf{92.00} &\textbf{95.88} \\
\hline
\end{tabular}
\end{center}
\end{table}

\subsection{Pedestrian Re-identification}
\begin{figure}[ht]
\centering
 {\includegraphics[width=0.20 \textwidth]{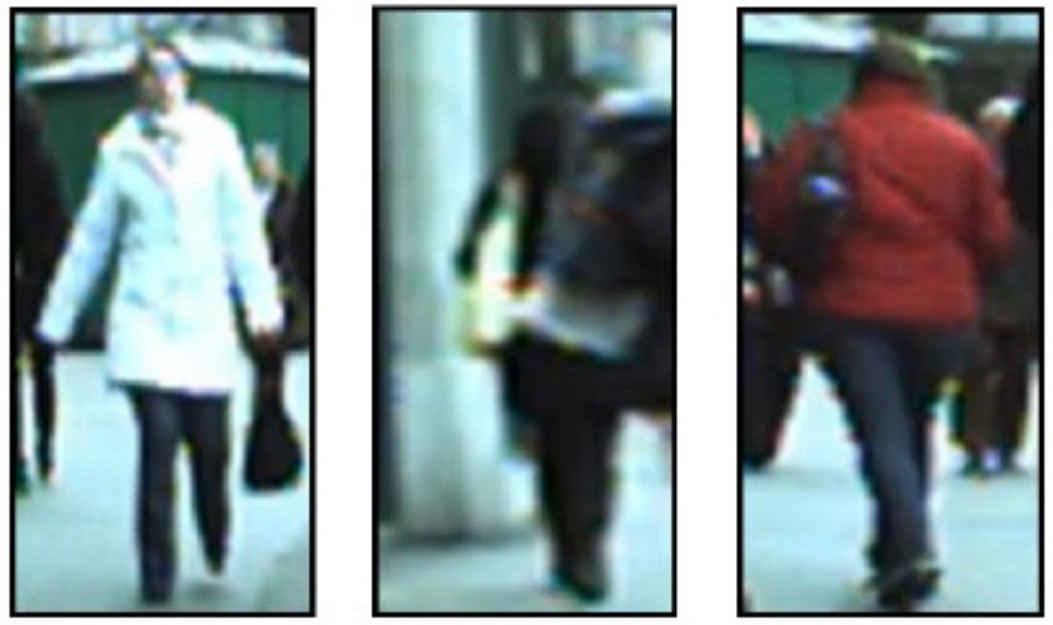}}  %
 {\includegraphics[width=0.20 \textwidth]{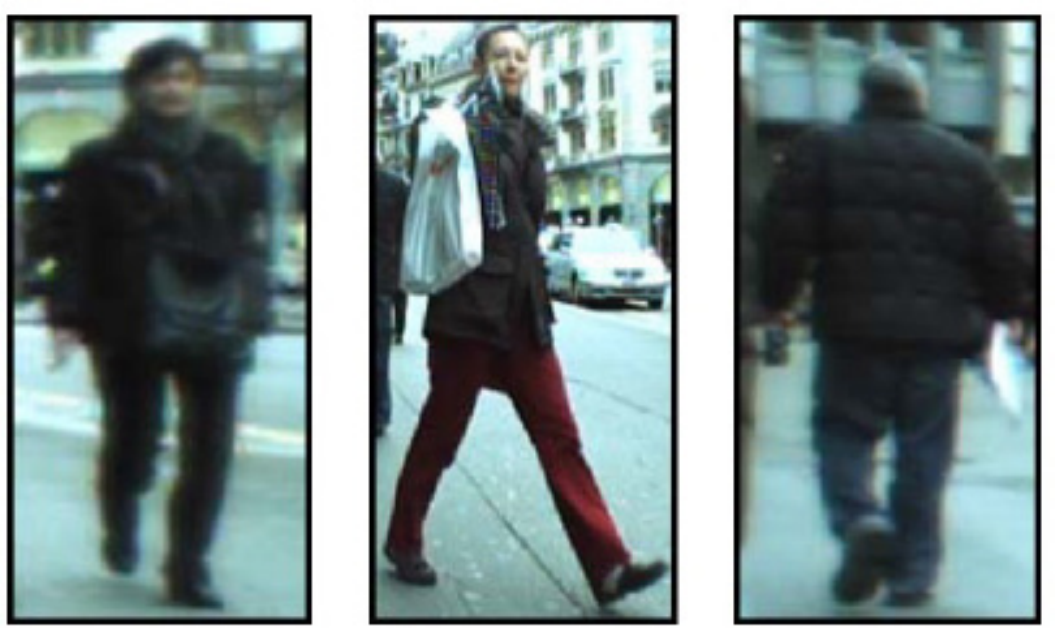}}
   \caption{Samples from the ETHZ dataset\citep{HarandiSandersonHartleyLovell2012}. }\label{fig4}
\end{figure}

Finally, we conduct the person re-identification task by our proposed method and compare with other methods. Here, we used the modified ETHZ dataset\citep{SchwartzDavis2009}, illustrated in Fig.\ref{fig4}. The original ETHZ includes 3 Sequences, in which Sequence 1 contains 83 pedestrians (4,857 images), Sequence 2 contains 35 pedestrians (1,936 images), and Sequence 3 contains 28 pedestrians (1,762 images). To facilitate the subsequent processing, we first down-sampled all images to 64$\times$32 pixels following the work\citep{HarandiHartleyLovellSanderson2014}. To prepare the covariance descriptors, the following features are utilized: the position of pixel, the color information from RGB channels, the gradient and Laplacian information from the corresponding color part, respectively. That is, each region can be depicted by a 17$\times$17 covariance matrix. To constructing the training samples, 10 images are randomly selected from each subject while the rest are used for testing. For fairly comparison, we adopt five splits for  each sequence to test the classification performance.

The recognition results are presented in Tables\ref{Tab5}-\ref{Tab7}. We tuned the parameters for each method to achieve the best results and reported them in tables. And the best results for each test Seq. are highlighted in bold numbers as usual. For the methods using kernel trick, the second parameter in the brackets denotes the kernel parameter. As can be seen, the proposed NPKSRC achieves the best score for each sequence in average sense. While for LogE-GkSRC, it obtains the second best results in terms of classification rate thanks to the use of Log-Euclidean metric. To explain this observation, it may owe to considering the weight structure within data. As for RSRS, it applies the Stein kernel inferior to the methods using Log-Euclidean Gaussian one.

\begin{figure}[ht]
\centering
 {\includegraphics[width=0.50 \textwidth]{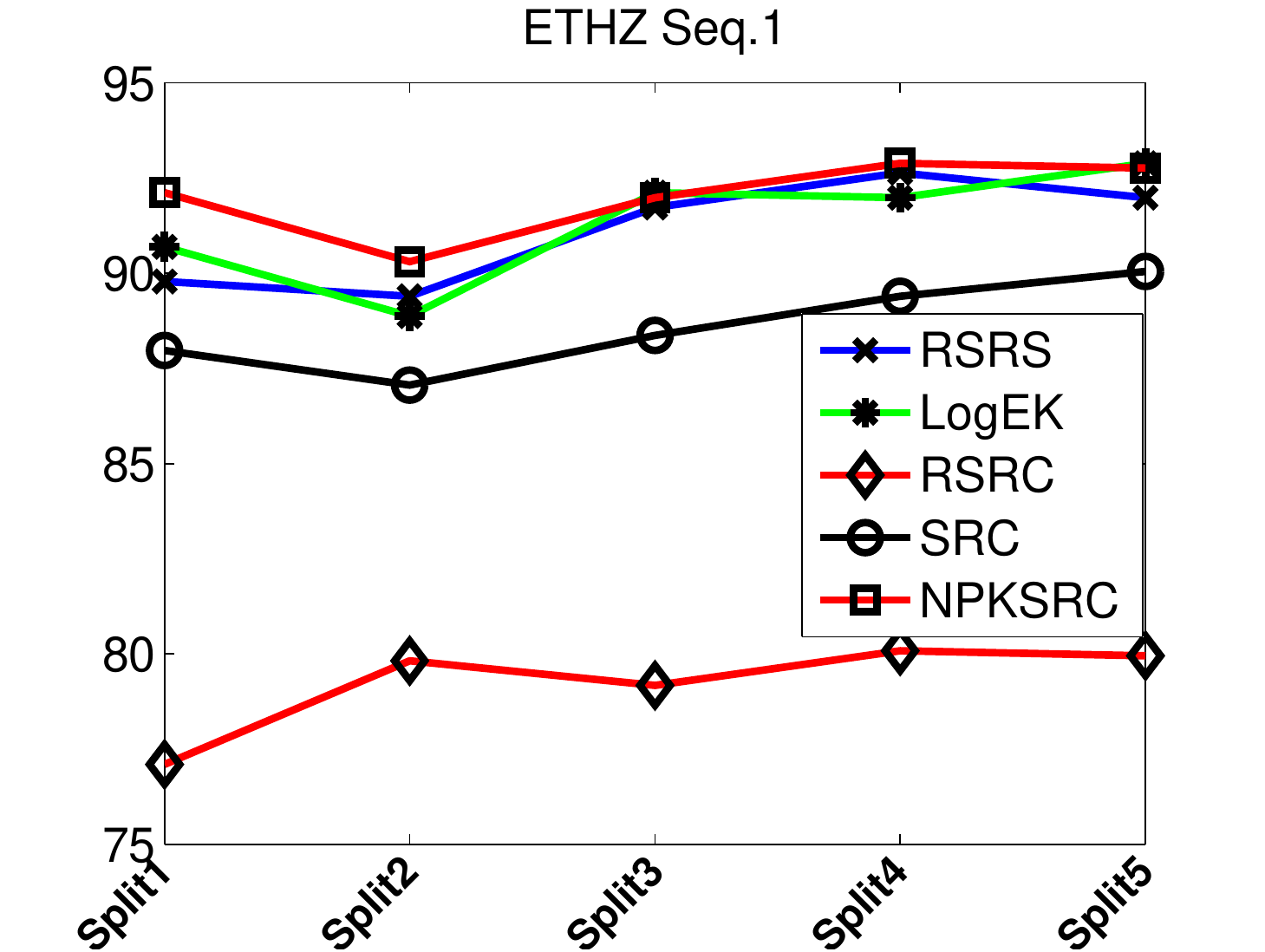}}  % {\includegraphics[width=0.20 \textwidth]{ETHZ_p2.pdf}}
   \caption{Recognition results from the ETHZ dataset. }\label{fig5}
\end{figure}

Furthermore, to clearly show the advantage of our method, we plot a recognition rate vs. each split for seq.1 in Fig.\ref{fig5}. As the curves for seq.2 and seq.3 are similar to that of seq.1, we do not repeatedly present here.

\begin{table*}
\caption{Classification results in terms of accuracy (\%) on ETHZ Seq.1 dataset.}\label{Tab5}
\begin{center}
\begin{tabular}{|l|c|c|c|c|c|c| }
\hline
$Dataset$ & $s_1$ & $s_2$ & $s_3$ &$s_4$ & $s_5$ & avg.\\
\hline\hline
SRC\citep{WrightYangGaneshSastryMa2009}(0.001) &87.97 &87.06 &88.36	&89.39	&90.04	&88.56 \\
RSRC\citep{CherianSra2014}(0.01) &77.10 &79.82	&79.17 &80.08 &79.95 &79.22  \\
RSRS\citep{HarandiHartleyLovellSanderson2014}(0.01,10) & 89.78 &89.39	&91.72 &92.63 &91.98 &91.10    \\
LogE-GkSRC\citep{LiWangZuoZhang2013}(0.001,0.02)  &90.69	&88.87	&\textbf{92.11}	&91.98 &92.88 &91.31 \\
NPKSRC (0.001,0.001) &\textbf{92.11} &\textbf{90.30} &91.98 &\textbf{92.88} &\textbf{92.76}	&\textbf{92.00}  \\
\hline
\end{tabular}
\end{center}
\end{table*}

\begin{table*}
\caption{Classification results in terms of accuracy (\%) on ETHZ Seq.2 dataset.}\label{Tab6}
\begin{center}
\begin{tabular}{|l|c|c|c|c|c|c| }
\hline
$Dataset$ & $s_1$ & $s_2$ & $s_3$ &$s_4$ & $s_5$ & avg.\\
\hline\hline
SRC\citep{WrightYangGaneshSastryMa2009}(0.001) &83.74 &85.28 &85.89	&86.81 &86.50 &85.64  \\
RSRC\citep{CherianSra2014}(0.01) &84.66 &81.60 &82.52 &86.20 &83.44 &83.68\\
RSRS\citep{HarandiHartleyLovellSanderson2014}(0.01,10) &90.49 &88.96 &\textbf{89.88} &90.80 &\textbf{91.41} &90.31 \\
LogE-GkSRC\citep{LiWangZuoZhang2013}(0.001,0.02)  &91.41 &89.88 &88.65 &92.94 &90.80 &90.74 \\
NPKSRC (0.001,0.001) &\textbf{91.72} &\textbf{89.88} &88.35 &\textbf{93.25} & 90.49 &\textbf{90.74} \\
\hline
\end{tabular}
\end{center}
\end{table*}

\begin{table*}
\caption{Classification results in terms of accuracy (\%) on ETHZ Seq.3 dataset.}\label{Tab7}
\begin{center}
\begin{tabular}{|l|c|c|c|c|c|c| }
\hline
$Dataset$ & $s_1$ & $s_2$ & $s_3$ &$s_4$ & $s_5$ & avg.\\
\hline\hline
SRC\citep{WrightYangGaneshSastryMa2009}(0.001) &95.92 &95.10 &90.61 &93.88 &93.06 &93.71\\
RSRC\citep{CherianSra2014}(0.01) &92.65 &92.65 &91.02 &91.02 &91.84 &91.84  \\
RSRS\citep{HarandiHartleyLovellSanderson2014}(0.01,10) &98.78	&97.55	&\textbf{98.37}	&96.73 &98.37 &97.96  \\
LogE-GkSRC\citep{LiWangZuoZhang2013}(0.001,0.02)  &\textbf{99.59}	&98.37 &97.55 &95.92 &98.37 &97.96  \\
NPKSRC (0.001,0.001) &99.18	&\textbf{99.59} &97.14 &\textbf{97.14} &\textbf{98.37} &\textbf{98.28}   \\
\hline
\end{tabular}
\end{center}
\end{table*}

\section{Conclusion}\label{SectionV}
In this paper, a novel robust classification algorithm, termed as neighborhood preserved sparse representation, is proposed for SPD matrices by fully exploiting the Riemannian geometry structure within data. Specifically, the local consistency constraint, formulated by the geodesic distance under Log-Euclidean metric, is imposed onto the sparse coding paradigm over a Riemannian manifold. Experimental results show that the proposed method can provide better classification solutions than the state-of-the-art approaches thanks to incorporating Riemannian geometry structure.

Although our proposed method achieved promising performance in terms of recognition rate, there still exists some open issues deserving to study. One prompt direction may be how to devise a better weighted constraint such that the beneficial discriminant representations can be achieved.

\section{Acknowledgement}
The Project was supported in part by the Guangdong Natural Science Foundation under Grant (No. 2014A030313511) and in part by the Scientific Research Foundation for the Returned Overseas
Chinese Scholars, State Education Ministry, China.

\section*{Appendix}
Given a least-squares problem as following,
\begin{align}
\mathop{\min}\limits_{\mathbf{c}} \left\| \phi(X)-  \sum\limits_{i = 1}^N {c}_{i}\phi(Y_i)  \right\|_2^2.\label{ProbSPD}
\end{align}
where data $\mathcal{Y}= [Y_1,Y_2,..., Y_N]$ and $X$ are on SPD manifold $\mathcal{S}_d^+$. This problem can be rewritten as a least-squares problem on Euclidean space. That is,
\begin{align}
\mathop{\min}\limits_{\mathbf{c}}  \|\mathbf{\bar{x}}- \bar{D} \mathbf{c}\|_2^2.\label{ProbSPD2}
\end{align}
where $\mathbf{\bar{x}}= \Sigma^{-1/2}U^T \kappa(X, \mathcal{Y})$ and $\bar{D} = \Sigma^{-1/2}U^T$, given the \text{SVD} of $\kappa(\mathcal{Y}, \mathcal{Y})$ is $U\Sigma U^T, UU^T = \textbf{I}$.

\begin{proof} By expanding the $\ell_2$-norm term in problem \eqref{ProbSPD}, we have the following formulation,
\begin{align}
&\mathop{\min}\limits_{\mathbf{c}}  \left\| \phi(X)-  \sum\limits_{i = 1}^N {c}_{i}\phi(Y_i)  \right\|_2^2 \notag \\
&= \mathop{\min}\limits_{\mathbf{c}}  \mathbf{c}^T \kappa(\mathcal{Y}, \mathcal{Y})\mathbf{c} - 2 \mathbf{c}^T \kappa(X, \mathcal{Y}) + f(X) \notag  \\
&= \mathop{\min}\limits_{\mathbf{c}}  \mathbf{c}^TU\Sigma U^T \mathbf{c}- 2\mathbf{c}^TU\Sigma^{-1/2}\Sigma^{1/2} U^T \kappa(X, \mathcal{Y}) \notag  \\ &+\kappa(X, \mathcal{Y})^T U\Sigma^{-1/2}\Sigma^{-1/2} U^T \kappa(X, \mathcal{Y}) \notag  \\
&=\mathop{\min}\limits_{\mathbf{c}} \| \Sigma^{-1/2} U^T \kappa(X, \mathcal{Y})- \Sigma^{-1/2} U^T \mathbf{c} \|_2^2.
\end{align}
Let $\mathbf{\bar{x}}= \Sigma^{-1/2}U^T \kappa(X, \mathcal{Y})$ and $\bar{D} = \Sigma^{-1/2}U^T$, then the formulation \eqref{ProbSPD2} is recognized.
\end{proof}

%\bibliographystyle{plain}
%\bibliography{JabRef_Ming_Laptop.bib}

\begin{thebibliography}{10}

\bibitem{ArsignyFillardPennecAyache2007}
Vincent Arsigny, Pierre Fillard, Xavier Pennec, and Nicholas Ayache.
\newblock Geometric means in a novel vector space structure on symmetric
  positive-definite matrices.
\newblock {\em {SIAM} Journal on Matrix Analysis and Applications},
  29(1):328--347, 2007.

\bibitem{BoydParikhChuPeleatoEckstein2011}
S.~Boyd, N.~Parikh, E.~Chu, B.~Peleato, and J.~Eckstein.
\newblock {\em Distributed Optimization and Statistical Learning via the
  Alternating Direction Method of Multipliers}, volume~3.
\newblock Foundations and Trends in Machine Learning, 2011.

\bibitem{CherianSra2014}
Anoop Cherian and Suvrit Sra.
\newblock Riemannian sparse coding for positive definite matrices.
\newblock In {\em Proceedings of ECCV}, volume 8691, pages 299--314. Springer
  International Publishing, 2014.

\bibitem{DonohoEladTemlyakov2006}
D.L. Donoho, M.~Elad, and V.~Temlyakov.
\newblock Stable recovery of sparse overcomplete representations in the
  presence of noise.
\newblock {\em IEEE Transactions on Information Theory}, 52(1):6--18, 2006.

\bibitem{FuGaoHongTien2015}
Yifan Fu, Junbin Gao, Xia Hong, and David Tien.
\newblock Low rank representation on {R}iemannian manifold of symmetric
  positive deffinite matrices.
\newblock In {\em Proceedings of SDM}, 2015, DOI:10.1137/1.9781611974010.36.

\bibitem{HarandiSalzmann2015}
M.~Harandi and M.~Salzmann.
\newblock Riemannian coding and dictionary learning: Kernels to the rescue.
\newblock In {\em Proceedings of CVPR}, pages 3926--3935, 2015.

\bibitem{HarandiSandersonHartleyLovell2012}
M~Harandi, C~Sanderson, R~Hartley, and B~Lovell.
\newblock Sparse coding and dictionary learning for symmetric positive definite
  matrices: A kernel approach.
\newblock In {\em Proceedings of ECCV}, pages 216--229, 2012.

\bibitem{HarandiHartleyLovellSanderson2014}
Mehrtash~Tafazzoli Harandi, Richard Hartley, Brian~C. Lovell, and Conrad
  Sanderson.
\newblock Sparse coding on symmetric positive definite manifolds using bregman
  divergences.
\newblock {\em IEEE Transactions on Neural Networks and Learning Systems},
  2015, DOI:10.1109/TNNLS.2014.2387383.

\bibitem{HoXieVemuri2013}
Jeffrey Ho, Yuchen Xie, and Baba~C. Vemuri.
\newblock On a nonlinear generalization of sparse coding and dictionary
  learning.
\newblock In {\em Proceedings of ICML}, volume~28, pages 1480--1488, 2013.

\bibitem{JayasumanaHartleySalzmannLiHarandi2013}
Sadeep Jayasumana, Richard Hartley, Mathieu Salzmann, Hongdong Li, and
  Mehrtash~Tafazzoli Harandi.
\newblock Kernel methods on the {R}iemannian manifold of symmetric positive
  definite matrices.
\newblock In {\em Proceedings of CVPR}, pages 73--80, June 2013.

\bibitem{LiWangZuoZhang2013}
Peihua Li, Qilong Wang, Wangmeng Zuo, and Lei Zhang.
\newblock Log-{E}uclidean kernels for sparse representation and dictionary
  learning.
\newblock In {\em Proceedings of ICCV}, pages 1601--1608, Dec 2013.

\bibitem{LinLiuLi2015}
Z.~Lin, R.~Liu, and H.~Li.
\newblock Linearized alternating direction method with parallel splitting and
  adaptive penalty for separable convex programs in machine learning.
\newblock {\em Machine Learning}, 99(2):287--325, 2015.

\bibitem{LinChenMa2009}
Zhouchen Lin, Minming Chen, and Yi~Ma.
\newblock The augmented lagrange multiplier method for exact recovery of
  corrupted low-rank matrices.
\newblock Technical report, UIUC Technical Report UILU-ENG-09-2215, 2009.

\bibitem{PennecFillardAyache2006}
Xavier Pennec, Pierre Fillard, and Nicholas Ayache.
\newblock A {R}iemannian framework for tensor computing.
\newblock {\em International Journal Of Computer Vision}, 66:41--66, 2006.

\bibitem{RoweisSaul2000}
S.~T. Roweis and L.~K. Saul.
\newblock Nonlinear dimensionality reduction by locally linear embedding.
\newblock {\em Science}, 290:2323--2326, 2000.

\bibitem{SchwartzDavis2009}
W.R. Schwartz and L.S. Davis.
\newblock Learning discriminative appearance-based models using partial least
  squares.
\newblock In {\em Proceedings of {SIBGRAPI}}, pages 322--329, 2009.

\bibitem{SivalingamBoleyMorellasPapanikolopoulos2014}
Ravishankar Sivalingam, Daniel Boley, Vassilios Morellas, and Nikolaos
  Papanikolopoulos.
\newblock Tensor sparse coding for positive definite matrices.
\newblock {\em IEEE Transactions on Pattern Analysis and Machine Intelligence},
  36(3):592--605, 2014.

\bibitem{TenenbaumSilvaLangford2000}
J.~B. Tenenbaum, V.~d.~Silva, and J.~C. Langford.
\newblock A global geometric framework for nonlinear dimensionality reduction.
\newblock {\em Science}, 290:2319--2323, 2000.

\bibitem{TuzelPorikliMeer2008}
O.~Tuzel, F.~Porikli, and P.~Meer.
\newblock Pedestrian detection via classification on {R}iemannian manifolds.
\newblock {\em IEEE Transactions on Pattern Analysis and Machine Intelligence},
  30(10):1713--1727, Oct 2008.

\bibitem{WangHuGaoSunYin2015}
B.Y. Wang, Y.L. Hu, J.~Gao, Y.F. Sun, and B.C. Yin.
\newblock Low rank representation on {G}rassmann manifolds: An extrinsic
  perspective.
\newblock {\em arXiv preprint arXiv:1504.01807}.

\bibitem{WangYangYuLvHuangGong2010}
Jinjun Wang, Jianchao Yang, Kai Yu, Fengjun Lv, Thomas Huang, and Yihong Gong.
\newblock Locality-constrained linear coding for image classification.
\newblock In {\em Proceedings of CVPR}, 2010.

\bibitem{WangShanChenGao2008}
Ruiping Wang, Shiguang Shan, Xilin Chen, and Wen Gao.
\newblock Manifold-manifold distance with application to face recognition based
  on image set.
\newblock In {\em Proceedings of CVPR}, pages 1--8, June 2008.

\bibitem{WangSlavakisLerman2015}
Xu~Wang, Konstantinos Slavakis, and Gilad Lerman.
\newblock Multi-manifold modeling in non-euclidean spaces.
\newblock In {\em Proceedings of {AISTATS} 2015, San Diego, California, USA,
  May 9-12, 2015}, 2015.

\bibitem{WrightYangGaneshSastryMa2009}
John Wright, Allen~Y. Yang, Arvind Ganesh, S.~Shankar Sastry, and Yi~Ma.
\newblock Robust face recognition via sparse representation.
\newblock {\em IEEE Transactions on Pattern Analysis and Machince
  Intelligence}, 31(2):210--227, February 2009.

\bibitem{YangWrightMaSastry2008}
Allen~Y. Yang, John Wright, Yi~Ma, and Shankar Sastry.
\newblock Unsupervised segmentation of natural images via lossy data
  compression.
\newblock {\em Computer Vision and Image Understanding}, 110(2):212--225, 2008.

\bibitem{YangChuZhangXuYang2013}
Jian Yang, Delin Chu, Lei Zhang, Yong Xu, and Jingyu Yang.
\newblock Sparse representation classifier steered discriminative projection
  with applications to face recognition.
\newblock {\em IEEE Transactions on Neural Networks and Learning Systems},
  24(7):1023--1035, July 2013.

\bibitem{YangZhangShiuZhang2013}
Meng Yang, Lei Zhang, Simon~C.K. Shiu, and David Zhang.
\newblock Gabor feature based robust representation and classification for face
  recognition with {G}abor occlusion dictionary.
\newblock {\em Pattern Recognition}, 46(7):1865 -- 1878, 2013.

\bibitem{YinLiuJinYang2012}
Jun Yin, Zhonghua Liu, Zhong Jin, and Wankou Yang.
\newblock Kernel sparse representation based classification.
\newblock {\em Neurocomputing}, 77(1):120 -- 128, 2012.

\bibitem{YinGaoGuo2015}
Ming Yin, Junbin Gao, and Yi~Guo.
\newblock Nonlinear low-rank representation on {S}tiefel manifolds.
\newblock {\em Electronics Letters}, 51(10):749--751, 2015.

\bibitem{YinGaoTienCai2014}
Ming Yin, Junbin Gao, David Tien, and Shuting Cai.
\newblock Blind image deblurring via coupled sparse representation.
\newblock {\em Journal of Visual Communication and Image Representation},
  25(5):814--821, July 2014.

\bibitem{ZhangZhouChangLiuYanWangLi2012}
Li~Zhang, Wei-Da Zhou, Pei-Chann Chang, Jing Liu, Zhe Yan, Ting Wang, and
  Fan-Zhang Li.
\newblock Kernel sparse representation-based classifier.
\newblock {\em IEEE Transactions on Signal Processing}, 60(4):1684--1695, 2012.

\end{thebibliography}

\end{document}